# A personalized Uncertainty Quantification framework for patient survival models: estimating individual uncertainty of patients with metastatic brain tumors in the absence of ground truth

Yuqi Wang, Aarzu Gupta, David Carpenter, Trey Mullikin, Zachary J. Reitman, Scott Floyd, John Kirkpatrick, Joseph K. Salama, Paul W. Sperduto, Jian-Guo Liu, Mustafa R. Bashir, Kyle J. Lafata

***Abstract*—** To develop a novel Uncertainty Quantification (UQ) framework to estimate the uncertainty of patient survival models in the absence of ground truth, we developed and evaluated our approach based on a dataset of 1383 patients treated with stereotactic radiosurgery (SRS) for brain metastases between January 2015 and December 2020. Our motivating hypothesis is that a time-to-event prediction of a test patient on inference is more certain given a higher feature-space-similarity to patients in the training set. Therefore, the uncertainty for a particular patient-of-interest is represented by the concordance index between a patient similarity rank and a prediction similarity rank. Model uncertainty was defined as the increased percentage of the max uncertainty-constrained-AUC compared to the model AUC. We evaluated our method on multiple clinically-relevant endpoints, including time to intracranial progression (ICP), progression free survival (PFS) after SRS, overall survival (OS), and time to ICP and/or death (ICPD), on a variety of both statistical and non-statistical models, including CoxPH, conditional survival forest (CSF), and neural multi-task linear regression (NMTLR). Our results show that all models had the lowest uncertainty on ICP (2.21%) and the highest uncertainty (17.28%) on ICPD. OS models demonstrated high variation in uncertainty performance, where NMTLR had the lowest uncertainty (1.96%) and CSF had the highest uncertainty (14.29%). In conclusion, our method can estimate the uncertainty of individual patient survival modeling results. As expected, our data empirically demonstrate that as model uncertainty measured via our technique increases, the similarity between a feature-space and its predicted outcome decreases.

## I. INTRODUCTION

TIME-TO-EVENT analysis has wide applications in various domains, such as healthcare [1]–[3], finance [4]–[7], social science [8]–[12], etc. One crucial application in healthcare is survival analysis, where statistical methods are used to analyze and interpret the time duration until an event of interest occurs. Such analyses are essential to interpreting efficacy of new treatments and may help healthcare professionals to make informed decisions for patient care. For example, over the past 25 years, oncologists have relied on estimates of overall survival (OS) and intracranial metastatic progression (ICP) to guide management decisions for patients with brain metastases [13], [14]. Such estimations are common in clinical practice, where brain metastases develop in approximately 20-40% of all cancer patients [15]. While median survival rates are generally poor (< 12 months), significant heterogeneity is observed across clinical outcomes including death and ICP [15], [16]. Accordingly, prognostic models with respect to both OS [17], [18] and ICP [19], [20] have become an essential component of judicious selection across therapeutic approaches that span surgery, systemic therapy, and radiation therapy. Improved prognostication may optimize both patient-level clinical decision support and population-level clinical trial design.

The primary goal of survival analysis is to estimate the probability of occurrence of the event of interest over time and to investigate the impact of various covariates on the event probability. Therefore, it can be modeled as either a regression

Corresponding author: Kyle Lafata.

Yuqi Wang and Aarzu Gupta is with the Department of Electrical and Computer Engineering, Duke University, Durham, NC 27705 USA (e-mails: yuqi.wang@duke.edu, and aarzu.gupta@duke.edu).

David Carpenter, Trey Mullikin, Zachary Reitman, Scott Floyd, John Kirkpatrick, and Joseph Salama are with the Department of Radiation Oncology, Duke University, Durham, NC 27705 USA (e-mail: david.carpenter@duke.edu, trey.mullikin@duke.edu, zachary.reitman @duke.edu, scott.floyd@duke.edu, john.kirkpatrick@duke.edu, and joseph.salama@duke.edu).

Jian-guo Liu is with the Department of Mathematics and Department of Physics, Duke University, Durham, NC 27705 USA (e-mail: jian-guo.liu@duke.edu).

Mustafa Bashir is with the Department of Radiology, the Department of Medicine, and the Center for Advanced Magnetic Resonance Development, Duke University, Durham, NC 27705 USA (e-mail: mustafa.bashir@duke.edu).

Kyle Lafata is with the Department of Electrical and Computer Engineering, the Department of Radiation Oncology, and the Department of Radiology.

problem, a classification problem, or a multi-task problem. In the regression approach, the focus is on predicting the time to the event of interest. The goal is to estimate the hazard function or survival function, which describes the probability of an event occurring at a specific time given the relevant covariates. In contrast, the classification approach aims to predict whether an event will occur within a given time window, without explicitly estimating the exact time to the event. In this case, the response variable is binary, indicating the occurrence or non-occurrence of the event. Combining both ideas, it can also be modeled as a multi-task problem, predicting whether and when the event will happen simultaneously.

In the past few decades, there has been a rapid development of various survival analysis models and techniques, including both statistical and non-statistical methods [21]–[28]. Commonly used statistical methods include Cox proportional hazards (CoxPH) models [27], which assumes that the hazard function is proportional to a linear combination of covariates, and parametric survival models, which specify a functional form of the hazard function. In contrast, non-statistical survival analysis methods employ machine learning algorithms such as random forests [26], support vector machines [28], and neural networks [21]–[24] to predict the time-to-event outcomes.

Unfortunately, the application of survival analysis models to real-world problems often presents several challenges characterized by uncertainty. While statistical methods are generally preferred for their interpretability and ability to handle missing data, non-statistical methods have shown promise in handling complex relationships between predictors and outcomes and may outperform statistical methods in certain cases. Therefore, it is often challenging to choose an optimal model. A related challenge is the uncertainty caused by censored data, which refers to observations where the event of interest has not yet occurred at the study endpoint, or where a patient is lost to follow-up. More specifically, right-censoring occurs when the event of interest has not yet occurred by the end of the study, or the patient is lost to follow-up before the event occurs. Meanwhile, left-censoring occurs when the event of interest occurred before the study's start, and only the time to the event is unknown. A variety of statistical techniques have been developed to handle this type of data, but a large proportion of censored observations will still decrease model generalization and lead to higher uncertainty of prediction due to the incomplete information.

Importantly, the event threshold of current state of the art survival models is generally decided by the model performance on the test set as a whole, without uncertainty quantification (UQ) on a personalized individual patient basis. However, estimating the uncertainty of these models is challenging due to the lack of ground-truth in survival analysis. Therefore, UQ methods are needed to analyze the uncertainty of individual patient predictions in the absence of ground truth, which is the clinically relevant, real-world scenario for prospectively monitoring patients throughout their care.

To address this issue, researchers have proposed various probabilistic approaches for UQ in survival analysis, which can be divided into two categories: non-sampling methods and Monte Carlo sampling methods [29]. For non-sampling method, one such approach is the use of Bayesian methods to estimate posterior distributions of survival probabilities. For example, García-Donato et al. used Cox regression, a semi-parametric approach, from a probabilistic perspective within a Bayesian framework, deriving a new prior distribution for the specific model parameters. This method permits treating the risk function as an unknown, reducing the need for additional assumptions. However, it cannot be applied to other parametric survival models and there is no personalized UQ score for each case [30].

Additionally, there have been efforts to incorporate various parametric techniques into Monte Carlo sampling UQ methods in survival analysis. For instance, Loya et al. developed a deep learning-based Bayesian framework that simultaneously pre- dicted and quantified uncertainty by weight sampling, which accurately identified the out-of-distribution cohort. Unfortu- nately, this method is unable to provide a UQ score for distinct models or individual cases [31]. Sokota et al. employed a personalized uncertainty representation technique by acquiring the confidence intervals of the pointwise prediction utilizing MCMC weight-sampled models. However, this approach was only effective for parametric models as well and cannot provide an overall model uncertainty score [32].

To fill the gap of model-independent UQ method, in this paper, we propose a novel UQ framework for survival analysis that is capable of estimating model uncertainty for both semi-parametric and parametric models based on a patient-specific UQ score. Our approach is based on the hypothesis that the prediction of a test patient on inference is more certain given a higher feature-space-similarity to patients in the training set. Therefore, we propose to represent the uncertainty for a particular patient-of-interest by the concordance index between a patient similarity rank and a prediction similarity rank, and the uncertainty for a model as the increased model AUC after introducing the personalized UQ score. Specifically in the brain metastases UQ problem, we define the patient similarity loss as the sum-of-the-difference between a clinical nomogram of ICP and the feature-level 0-1 loss, and prediction similarity loss as squared loss. Then we cluster patients into groups and calculate the group-level concordance index to reduce the effect of noise within the outcomes data.

To the best of our knowledge, this is the first UQ method that incorporates both patient similarity and prediction similarity to estimate the uncertainty of survival models and the first model independent UQ method in survival analysis. We evaluate our approach on a dataset of 1383 patients treated with stereotactic radiosurgery for brain metastases and compare it against various traditional model evaluation metrics, such as concordance index (C-index) and integrated

Brier score (IBS). We also compare our approach with various statistical and non-statistical models, including CoxPH [27], conditional survival forest (CSF) [26], and neural multi-task linear regression (NMTLR) [21].

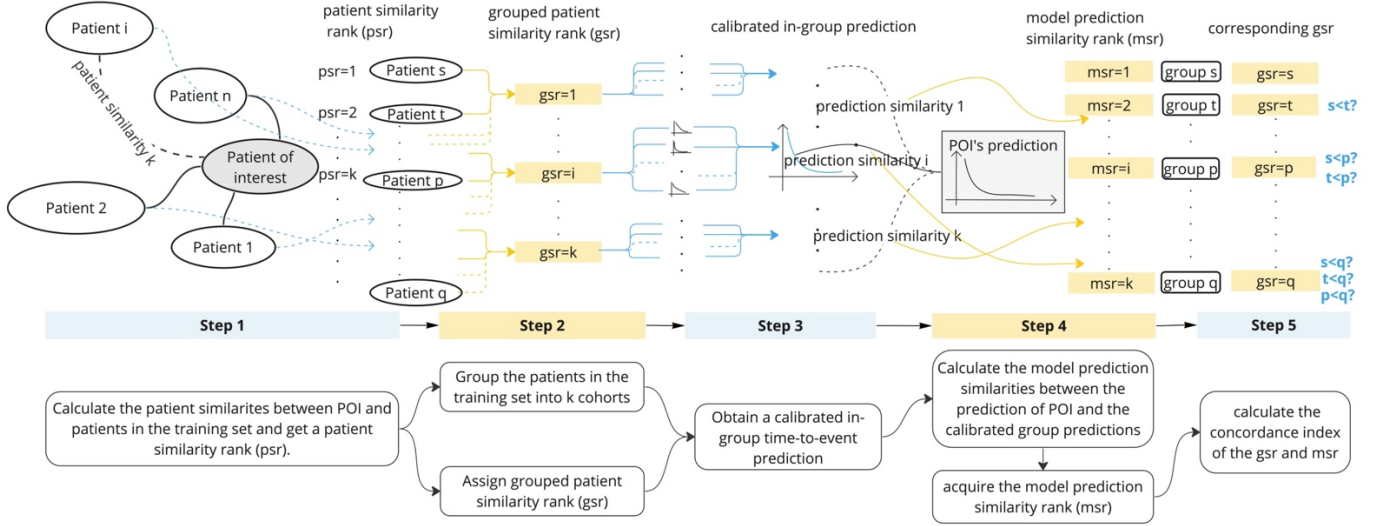

Fig. 1. **Overall workflow of the proposed personalized uncertainty quantification framework. Step1**: Given a patient of interest (POI), patient similarity is calculated by comparing the POI to the remaining patients in the training set. We define the loss as the sum-of-the-difference between a clinically relevant nomogram of intra-cranial progression and the feature-level 0-1 loss. Patients are then ranked based on their similarity score. **Step 2**: Patients in the training set are clustered based on their similarity loss (Step 1) to determine a group-level patient similarity rank. **Step 3**: Group-level calibrated model predictions are obtained by calculating the weighted average of all predictions within the group. **Step 4**: Prediction similarity is measured by calculating the squared loss between each in-group calibrated prediction and the prediction of the POI. The groups are then ranked based on model prediction loss. **Step 5**: A personalized Uncertainty Quantification score is defined as the concordance index of the group similarity rank (Step 2) and the model prediction similarity rank (Step 4).

## II. MATERIALS AND METHODS

### A. Dataset

We developed our method on a dataset of 1383 patients treated with stereotactic radiosurgery (SRS) for brain metastases between January 2015 and December 2020 [20]. The dataset contains 32 features, including demographics, molecular variants, pre-SRS disease and treatments, SRS treatment parameters, and post-SRS systemic therapies. Regarding out- comes data, we considered 4 different types of events: (i) intracranial progression (ICP), (ii) progression free survival (PFS), (iii) overall survival (OS), and (iv) ICP and/or death (ICPD). During the observation period, the uncensored rates of ICP, PFS, OS, and ICPD were 50.47%, 73.46%, 72.31%, and 87.20%, respectively. All research was conducted in keeping with best clinical practice and regulatory compliance. All data were generated from Pro00108434 approved by the Duke Institutional Review Board, which was completed under an approved waiver of consent, Health Insurance Portability and Protection Act Authorization, and decedent research notification.

### B. Personalized Uncertainty Quantification Framework

Fig. 1 illustrates our entire methodological workflow for personalized UQ applied to survival analyses. The process is briefly described as follows. Given a particular patient of interest (POI), we first calculate the similarity between each patient in the training set and the POI as shown in Step 1 of Fig. 1. Based on this similarity measurement, patients in the training set are clustered into k different groups as illustrated in Step 2 of Fig. 1. The clusters are utilized to obtain a group-level calibrated model prediction, which we use to calculate the prediction similarity between the POI and the calibrated in- group subset of patients as shown in Step 3 of Fig. 1. Finally, the personalized UQ score associated with the POI is defined as the concordance index between patient-level similarities and model-prediction-level similarities as illustrated in Step 4-5 of Fig. 1.

### C. Patient Similarity Loss and Model Prediction Similarity Loss

In this work, patient-level similarity was defined based on a clinical ICP nomogram score as shown in Table I. However, a limitation of this nomogram is that closer scores may repre- sent different feature combinations, and even the same score could represent different situations. For example, patients with different recurrence rates at 1 year and 5 years will receive the same score. Therefore, we modified the nomogram by adding an entry-level zero-one loss term. Thus, the loss function of patient similarity for the patient of interest POI and any patient p in the training set P is defined in equation (1),

$$L_{patient}(p, POI) = L_{nomogram}(p, POI) + L_{entry}(p, POI), \quad (1)$$

equation (2),

$$L_{nomogram}(p, POI) = |nomogram(POI) - nomogram(p)|, \quad (2)$$

and equation (3),

$$L_{entry}(p, POI) = ||POI \oplus p||_0. \quad (3)$$

Next, we sort p within P in increasing order of $L_{patient}$ and define the patient similarity rank psr for p as the sorted index. Equations (1), (2), and (3) mathematically described Step 1 of Fig. 1. To reduce the effect of potential hazards that are not reflected within the data and the noise inside the data, instead of considering the similarity between individuals, as shown in Step 2 of Fig. 1, we cluster $P$ into k groups $G$ and assign the group patient similarity rank $gsr$ as,

$$(g|gsr(g)) = (\{P_{psr}, psr \in [(i-1)*size_g + 1, i*size_g]\}|i), i \leq k, i \in \mathbb{N}, \quad (4)$$

where $size_g$ is the average size of each group,

$$size_g = \frac{size(P)}{k}.$$

As for patient similarity loss, we define the model-prediction-level loss between $pred_g$ and $pred_{POI}$ as the prediction of each group and $POI$,

$$L_{pred}(pred_{POI}, pred_g) = ||pred_{POI} - pred_g||_2, \quad (5)$$

where $pred_g$, the prediction of each group is defined as the weighted average of all predictions in group,

$$pred_g = \sum_{p \in g} softmax\left(-L_{patient}(p, POI)\right) * pred_p.$$

With the in-group calibrated prediction defined above and illustrated in Step 3 of Fig. 1, the model-prediction similarity rank $msr$ shown in Step 4 of Fig. 1 is then acquired by assigning the index of the increasingly sorted $L_{pred}$.

TABLE I
CLINICAL INTRACRANIAL PROGRESSION NOMOGRAM

| Nomogram Criteria | Treated brain metastases (Melanoma) | Treated brain metastases (Non-Melanoma) | History of whole brain radiotherapy | Time from Cancer Diagnosis to Initial Metastases | Total points |
|---|---|---|---|---|---|
| | 1 or 2: 35 points | 1: 0 points | Yes: 0 points | >5 years: 0 points | 0–85 points: Low Risk |
| | ≥3: 100 points | ≥2: 45 points | No: 15 points | ≤5 years: 45 points | ≥86 points: High Risk |

### D. Personalized UQ Score and Model Uncertainty

The personalized UQ score shown in Step 5 of Fig. 1 is defined as the concordance index between the group patient similarity rank $sr$ and the model prediction similarity rank $msr$,

$$uq_{POI} = concordance\_index(gsr, msr) = Probability(msr_i < msr_j | gsr_i < gsr_j). \quad (6)$$

The illustrating toy example in Fig. 2 provides a conceptual schematic of how we describe model uncertainty. As shown in Fig. 2(a), when the time-to-event task is simplified as a binary classification problem by assigning different model thresholds, we can perform ROC analysis to describe the model performance. As shown in Fig. 2(b), we evaluate the uncertainty of the model by adding a new model-independent dimension to the problem (i.e., the personalized UQ score defined in equation 6) for each prediction case. By applying different UQ score thresholds, the model will produce different area under the ROC curve (AUC) values. Therefore, model uncertainty can be represented by the increased percentage of the maximum AUC constrained by the personalized UQ score, relative to the un-constrained model AUC,

$$uncertainty_{model} = \frac{\max(AUC(uq_{threshold}))}{AUC_{model}}. \quad (7)$$

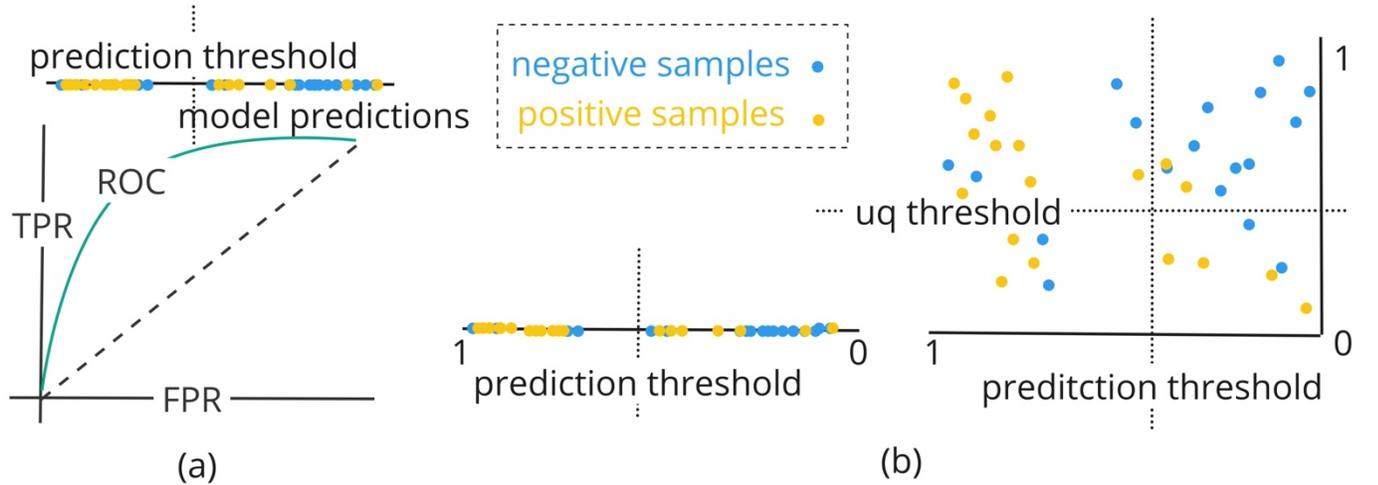

Fig. 2. A toy example to illustrate our defined model uncertainty.

## IV. Experiment and Results

### A. Experiment settings and model selections

To evaluate the effectiveness of our UQ algorithm, we tested it on various survival models, both statistical and non-statistical, and all three types of modeling. Specifically, we employed a Cox proportional hazard (CoxPH) model to represent statistical models, a conditional survival forest (CSF) to represent machine learning models, and a neural multi-task linear regression (NMTLR) to represent deep neural network models. Each of these models was applied to different time-to-event tasks, including ICP, PFS, OS, and ICPD. For our experiment, we calibrated our algorithm with k=10 groups. We utilized the nomogram from Table I and 32 features including demographics, molecular variants, pre-SRS disease and treatments, SRS treatment parameters, and post-SRS systemic therapies in our dataset. Patient similarity was calculated using equations (1), (2), and (3). Following clustering and ranking of groups using equation (4), model prediction similarity at the group level was determined using equations (5). The personalized UQ score was then calculated using equation (6). Applying this process to each patient, personalized UQ scores were obtained. Finally, model uncertainty was calculated using equation (7). We performed the process for the four endpoints, ICP, PFS, OS, and ICPD.

### B. Experiment results

The AUCs under different UQ constraints for different clinical endpoints are shown in Fig. 3. For each endpoint, all model AUCs increased as a function of the personalized UQ score threshold. More complete statistics are summarized in Table II, which shows the statistics of the traditional model evaluation metrics and our proposed UQ performance for different models applied to different time-to-event tasks. The time-dependent AUC (i.e., C-index), integrated Brier score (IBS), classification AUC, and maximum classification AUC are shown as model evaluation metrics, while the model uncertainty is reported as a measure of the normalized model performance on uncertainty estimation. The C-index is a measure of concordance between the predicted survival times and the observed survival times. It ranges from 0 to 1, with 0 indicating random prediction and 1 indicating perfect concordance between predicted and observed survival times. The IBS is a measure of the mean squared difference between the predicted probabilities of survival and the observed status over a given time interval. It ranges from 0 to 1, and a model with an IBS below 0.25 is considered informative.

Our results demonstrate that the NMTLR and CoxPH models have overall the best performance when evaluated using traditional model evaluation metrics (C-index and IBS). Specifically, CoxPH performed slightly better than NMTLR, especially on OS and ICPD, with a C-index of 0.72 for both events and an average IBS score of 0.11, whereas NMTLR has an average C-index of 0.64 and an average IBS score of 0.16. However, our proposed model uncertainty metric indicates that the CoxPH model is less certain than NMTLR for these events, with CoxPH exhibiting 6.38% model uncertainty on OS and 19.30% model uncertainty on ICPD, while NMTLR has lower model uncertainty at 1.96% and 15.00%, respectively. Furthermore, our results indicate that the CSF model performs poorly on all tasks based on both traditional model evaluation metrics and our proposed model uncertainty metric.

Comparing the model performance on different tasks, our results show that all models had the lowest average uncertainty on ICP (2.21%) and the highest average uncertainty (17.28%) on ICPD. OS models demonstrated high variation in uncertainty performance, where NMTLR had the lowest uncertainty (1.96%) and CSF had the highest uncertainty (14.29%).

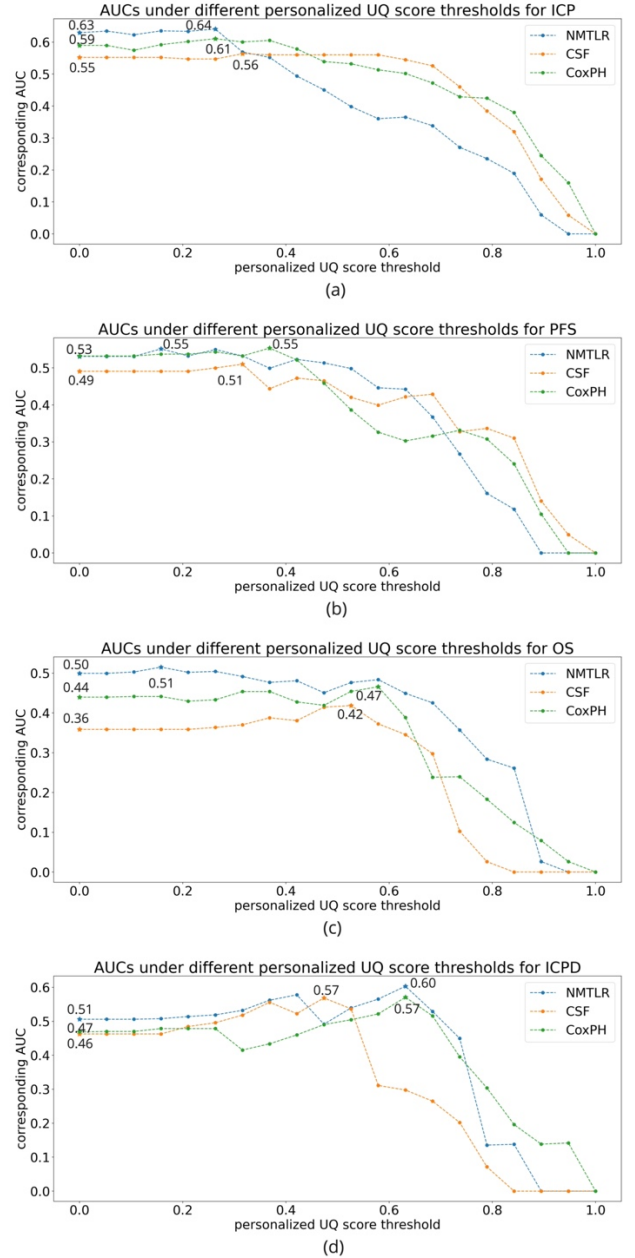

Fig. 3. AUCs under different warning certainty thresholds on different event predicting tasks. (a)(b)(c)(d) are the results on ICP, PFS, OS, and ICPD respectively.

TABLE II
STATISTICS OF THE MODEL EVALUATION METRICS AND UQ PERFORMANCE

| Hazards | ICP | | | PFS | | | OS | | | ICPD | | |
|---|---|---|---|---|---|---|---|---|---|---|---|---|
| Model/Metrics | C-index | IBS | Model uncertainty | C-index | IBS | Model uncertainty | C-index | IBS | Model uncertainty | C-index | IBS | Model uncertainty |
| NMTLR | 0.58 | 0.20 | 1.56% | 0.59 | 0.11 | 3.64% | 0.66 | 0.20 | 1.96% | 0.63 | 0.11 | 15.00% |
| CSF | 0.55 | 0.73 | 1.79% | 0.59 | 0.81 | 3.92% | 0.65 | 0.65 | 14.29% | 0.65 | 0.72 | 17.54% |
| CoxPH | 0.57 | 0.17 | 3.28% | 0.62 | 0.10 | 3.64% | 0.72 | 0.13 | 6.38% | 0.72 | 0.09 | 19.30% |

## V. Discussion

In this work, we proposed a personalized UQ algorithm that can measure uncertainty for individual patients when considering different clinical time-to-event problems. Our generalized technique was shown to have utility for both semi-parametric and parametric models. Our results demonstrate that all models have lowest uncertainty for ICP prediction, highest uncertainty for ICPD prediction, and second highest uncertainty for OS prediction. These findings are expected because the characteristics of the SRS data are highly relevant to ICP, while factors related to OS are generally of higher noise and therefore less likely to be captured with high certainty. Mixing two clinical endpoints, as our ICPD results demonstrate, introduces more variability into the models, which can be quantified by increased prediction uncertainty using our UQ method.

As for model selection, our analyses showed that model performance varied across different endpoints. The NMTLR model had the lowest uncertainty, while the CSF model had a relatively high uncertainty. While the CoxPH model was comparable to the NMTLR model in uncertainty for ICP and PFS, it demonstrated higher uncertainty for OS and ICPD. Importantly, our results suggest that model complexity is not always directly positively proportional to model uncertainty. For example, the neural network model (NMTLR) demonstrated lower uncertainty compared to the statistical model (CoxPH). This is likely related to increased data variability in the event space, which may impede some assumptions of traditional statistical methods.

Our proposed personalized UQ algorithm offers a solution for measuring uncertainty not only for individual patients, but also for semi-parametric and parametric models. While García-Donato et al. previously proposed a non-parametric model UQ method that quantifies CoxPH model uncertainty, it fails to provide personalized UQ scores [30]. In comparison, UQ methods for parametric survival analysis models, such as those proposed by Loya et al. [31] and Sokota et al. [32], quantify uncertainty by evaluating the variation of the predicting results of the Monte Carlo sampled model weights and can also provide a patient specific UQ score. However, these methods are limited to parametric models and require multiple modifications of the model weights during the UQ procedure. Furthermore, these methods are unable to give UQ scores for different models.

Our proposed UQ algorithm has several direct implications for clinical management of patients with brain metastases. First, variation in UQ estimates across clinical endpoints may inform oncologic trial design. As oncologic clinical trials fail to meet pre-specified primary endpoints at significantly higher rates than non-oncologic trials [33], several reports emphasize the need for improved methods of estimating the benefit of a given oncologic intervention[33], [34]. In the present study, the greater degree of model uncertainty observed across composite endpoints such as ICPD is particularly notable, as composite endpoints are now more commonly utilized than OS as a primary trial endpoint [35]. Moreover, as oncologic trials such as NRG Oncology BN009 increasingly integrate prognostic models of ICP and OS into patient selection [36], [37], UQ methods may similarly inform trial eligibility criteria. Second, the present UQ algorithm may offer valuable context for validation studies of established prognostic models. For example, a recent report using the present dataset validated the use of an ICP nomogram developed from an older patient cohort; while time to ICP was comparable across both datasets, OS was significantly greater in the more recent cohort [20], [38]. Finally, the UQ algorithm may assist clinical oncologists and patients with brain metastases in determining the degree to which prognostic models are applied within shared decision making. Overall, the proposed personalized UQ algorithm offers unique advantages to patient- and model-based clinical applicability.

While our findings demonstrate that our approach can estimate model uncertainty and can be used to improve the performance of survival models, there are still some limitations. First, the patient similarity metric we used is not a generalized metric, the ICP nomogram was part of the definition of patient similarity. Second, our study only used one dataset, which may also limit the generalizability of our findings. Third, this method requires a large amount of computational resource, as it requires comparison of the POI with all patients in the training set. Future studies are needed to validate our methods on additional datasets and to explore alternative techniques for more computationally efficient estimates of personalized and model uncertainty.

## VI. Conclusion

In summary, we proposed a personalized uncertainty quantification (UQ) algorithm that can measure uncertainty for both individual patients and the model. Our results show that the overall uncertainty is not only relevant to model selection, but also hazard: the greater the association between features and the predicted hazard, the lower the model uncertainty.

## References

[1] D. G. Beer *et al.*, "Gene-expression profiles predict survival of patients with lung adenocarcinoma," *Nat Med*, vol. 8, no. 8, pp. 816–824, Aug. 2002, doi: 10.1038/nm733.

[2] A. V. Antonov, M. Krestyaninova, R. A. Knight, I. Rodchenkov, G. Melino, and N. A. Barlev, "PPISURV:

a novel bioinformatics tool for uncovering the hidden role of specific genes in cancer survival outcome," *Oncogene*, vol. 33, no. 13, Art. no. 13, Mar. 2014, doi: 10.1038/onc.2013.119.

[3] D. Delen, G. Walker, and A. Kadam, "Predicting breast cancer survivability: a comparison of three data mining methods," *Artif. Intell. Med.*, vol. 34, no. 2, pp. 113–127, Jun. 2005, doi: 10.1016/j.artmed.2004.07.002.

[4] B. Baesens, S. Viaene, D. Van den Poel, J. Vanthienen, and G. Dedene, "Bayesian neural network learning for repeat purchase modelling in direct marketing," *European Journal of Operational Research*, vol. 138, no. 1, pp. 191–211, 2002.

[5] B. Baesens, T. V. Gestel, M. Stepanova, and D. V. D. Poel, "Neural Network Survival Analysis for Personal Loan Data," *Working Papers of Faculty of Economics and Business Administration, Ghent University, Belgium*, Art. no. 04/281, Nov. 2004, Accessed: Mar. 19, 2023. [Online]. Available: https://ideas.repec.org//p/rug/rugwps/04-281.html

[6] M. Stepanova and L. C. Thomas, "PHAB scores: proportional hazards analysis behavioural scores," *Journal of the Operational Research Society*, vol. 52, no. 9, pp. 1007–1016, 2001.

[7] M. Stepanova and L. Thomas, "Survival Analysis Methods for Personal Loan Data," *Operations Research*, vol. 50, no. 2, pp. 277–289, 2002.

[8] P. D. Berger and N. I. Nasr, "Customer lifetime value: Marketing models and applications," *Journal of Interactive Marketing*, vol. 12, no. 1, pp. 17–30, Jan. 1998, doi: 10.1002/(SICI)1520-6653(199824)12:1<17::AID-DIR3>3.0.CO;2-K.

[9] N. Barbieri, F. Silvestri, and M. Lalmas, "Improving Post-Click User Engagement on Native Ads via Survival Analysis," in *Proceedings of the 25th International Conference on World Wide Web*, in WWW '16. Republic and Canton of Geneva, CHE: International World Wide Web Conferences Steering Committee, Apr. 2016, pp. 761–770. doi: 10.1145/2872427.2883092.

[10] V. Rakesh, W.-C. Lee, and C. K. Reddy, "Probabilistic Group Recommendation Model for Crowdfunding Domains," *Proceedings of the Ninth ACM International Conference on Web Search and Data Mining*, pp. 257–266, Feb. 2016, doi: 10.1145/2835776.2835793.

[11] S. Ameri, M. J. Fard, R. B. Chinnam, and C. K. Reddy, "Survival Analysis based Framework for Early Prediction of Student Dropouts," in *Proceedings of the 25th ACM International on Conference on Information and Knowledge Management*, in CIKM '16. New York, NY, USA: Association for Computing Machinery, Oct. 2016, pp. 903–912. doi: 10.1145/2983323.2983351.

[12] P. A. Murtaugh, L. D. Burns, and J. Schuster, "PREDICTING THE RETENTION OF UNIVERSITY STUDENTS," *Research in Higher Education*, vol. 40, no. 3, pp. 355–371, Jun. 1999, doi: 10.1023/A:1018755201899.

[13] L. Gaspar et al., "Recursive partitioning analysis (RPA) of prognostic factors in three Radiation Therapy Oncology Group (RTOG) brain metastases trials," *Int J Radiat Oncol Biol Phys*, vol. 37, no. 4, pp. 745–751, Mar. 1997, doi: 10.1016/s0360-3016(96)00619-0.

[14] P. W. Sperduto, B. Berkey, L. E. Gaspar, M. Mehta, and W. Curran, "A new prognostic index and comparison to three other indices for patients with brain metastases: an analysis of 1,960 patients in the RTOG database," *Int J Radiat Oncol Biol Phys*, vol. 70, no. 2, pp. 510–514, Feb. 2008, doi: 10.1016/j.ijrobp.2007.06.074.

[15] D. N. Cagney et al., "Incidence and prognosis of patients with brain metastases at diagnosis of systemic malignancy: a population-based study," *Neuro Oncol*, vol. 19, no. 11, pp. 1511–1521, Oct. 2017, doi: 10.1093/neuonc/nox077.

[16] P. W. Sperduto et al., "Summary report on the graded prognostic assessment: an accurate and facile diagnosis-specific tool to estimate survival for patients with brain metastases," *J Clin Oncol*, vol. 30, no. 4, pp. 419–425, Feb. 2012, doi: 10.1200/JCO.2011.38.0527.

[17] M. Farris et al., "Brain Metastasis Velocity: A Novel Prognostic Metric Predictive of Overall Survival and Freedom From Whole-Brain Radiation Therapy After Distant Brain Failure Following Upfront Radiosurgery Alone," *Int J Radiat Oncol Biol Phys*, vol. 98, no. 1, pp. 131–141, May 2017, doi: 10.1016/j.ijrobp.2017.01.201.

[18] P. W. Sperduto et al., "Graded Prognostic Assessment (GPA) for Patients With Lung Cancer and Brain Metastases: Initial Report of the Small Cell Lung Cancer GPA and Update of the Non-Small Cell Lung Cancer GPA Including the Effect of Programmed Death Ligand 1 and Other Prognostic Factors," *Int J Radiat Oncol Biol Phys*, vol. 114, no. 1, pp. 60–74, Sep. 2022, doi: 10.1016/j.ijrobp.2022.03.020.

[19] D. N. Ayala-Peacock et al., "Prediction of new brain metastases after radiosurgery: validation and analysis of performance of a multi-institutional nomogram," *J Neurooncol*, vol. 135, no. 2, pp. 403–411, Nov. 2017, doi: 10.1007/s11060-017-2588-4.

[20] D. J. Carpenter et al., "Prognostic Model for Intracranial Progression after Stereotactic Radiosurgery: A Multicenter Validation Study," *Cancers*, vol. 14, no. 21, Art. no. 21, Jan. 2022, doi: 10.3390/cancers14215186.

[21] S. Fotso, "Deep Neural Networks for Survival Analysis Based on a Multi-Task Framework." arXiv, Jan. 16, 2018. Accessed: Jan. 13, 2023. [Online]. Available: http://arxiv.org/abs/1801.05512

[22] C. Lee, W. Zame, J. Yoon, and M. van der Schaar, "DeepHit: A Deep Learning Approach to Survival Analysis With Competing Risks," *Proceedings of the AAAI Conference on Artificial Intelligence*, vol. 32, no. 1, Art. no. 1, Apr. 2018, doi: 10.1609/aaai.v32i1.11842.

[23] C. Lee, J. Yoon, and M. van der Schaar, "Dynamic-DeepHit: A Deep Learning Approach for Dynamic Survival Analysis With Competing Risks Based on Longitudinal Data," *IEEE Transactions on Biomedical Engineering*, vol. 67, no. 1, pp. 122–133, Jan. 2020, doi: 10.1109/TBME.2019.2909027.

[24] J. Katzman, U. Shaham, J. Bates, A. Cloninger, T. Jiang, and Y. Kluger, "DeepSurv: Personalized

Treatment Recommender System Using A Cox Proportional Hazards Deep Neural Network," *BMC Med Res Methodol*, vol. 18, no. 1, p. 24, Dec. 2018, doi: 10.1186/s12874-018-0482-1.

[25] P. Wang, Y. Li, and C. K. Reddy, "Machine Learning for Survival Analysis: A Survey," *ACM Comput. Surv.*, vol. 51, no. 6, pp. 1–36, Nov. 2019, doi: 10.1145/3214306.

[26] M. N. Wright, T. Dankowski, and A. Ziegler, "Unbiased split variable selection for random survival forests using maximally selected rank statistics: M. WRIGHT, T. DANKOWSKI AND A. ZIEGLER," *Statist. Med.*, vol. 36, no. 8, pp. 1272–1284, Apr. 2017, doi: 10.1002/sim.7212.

[27] D. R. Cox, "Regression Models and Life-Tables," *Journal of the Royal Statistical Society: Series B (Methodological)*, vol. 34, no. 2, pp. 187–202, 1972, doi: 10.1111/j.2517-6161.1972.tb00899.x.

[28] S. Pölsterl, N. Navab, and A. Katouzian, "Fast Training of Support Vector Machines for Survival Analysis," in *Machine Learning and Knowledge Discovery in Databases*, A. Appice, P. P. Rodrigues, V. Santos Costa, J. Gama, A. Jorge, and C. Soares, Eds., in Lecture Notes in Computer Science. Cham: Springer International Publishing, 2015, pp. 243–259. doi: 10.1007/978-3-319-23525-7_15.

[29] J. Zhang, "Modern Monte Carlo methods for efficient uncertainty quantification and propagation: A survey," *WIREs Computational Statistics*, vol. 13, no. 5, p. e1539, 2021, doi: 10.1002/wics.1539.

[30] G. García-Donato, S. Cabras, and M. E. Castellanos, "Model uncertainty quantification in Cox regression," *Biometrics*, vol. n/a, no. n/a, doi: 10.1111/biom.13823.

[31] H. Loya, D. Anand, P. Poduval, N. Kumar, and A. Sethi, "A Bayesian framework to quantify survival uncertainty," *Annals of Oncology*, vol. 30, pp. vii32–vii33, Nov. 2019, doi: 10.1093/annonc/mdz413.116.

[32] S. Sokota, R. D'Orazio, K. Javed, H. Haider, and R. Greiner, "Simultaneous Prediction Intervals for Patient-Specific Survival Curves," *Proceedings of the Twenty-Eighth International Joint Conference on Artificial Intelligence*, pp. 5975–5981, Aug. 2019, doi: 10.24963/ijcai.2019/828.

[33] H. K. Gan, B. You, G. R. Pond, and E. X. Chen, "Assumptions of expected benefits in randomized phase III trials evaluating systemic treatments for cancer," *J Natl Cancer Inst*, vol. 104, no. 8, pp. 590–598, Apr. 2012, doi: 10.1093/jnci/djs141.

[34] L. Amiri-Kordestani and T. Fojo, "Why do phase III clinical trials in oncology fail so often?," *J Natl Cancer Inst*, vol. 104, no. 8, pp. 568–569, Apr. 2012, doi: 10.1093/jnci/djs180.

[35] J. C. Del Paggio *et al.*, "Evolution of the Randomized Clinical Trial in the Era of Precision Oncology," *JAMA Oncol*, vol. 7, no. 5, pp. 728–734, May 2021, doi: 10.1001/jamaoncol.2021.0379.

[36] V. Gondi, J. Meyer, and H. A. Shih, "Advances in radiotherapy for brain metastases," *Neurooncol Adv*, vol. 3, no. Suppl 5, pp. v26–v34, Nov. 2021, doi: 10.1093/noajnl/vdab126.

[37] E. R. McTyre *et al.*, "Multi-institutional validation of brain metastasis velocity, a recently defined predictor of outcomes following stereotactic radiosurgery," *Radiother Oncol*, vol. 142, pp. 168–174, Jan. 2020, doi: 10.1016/j.radonc.2019.08.011.

[38] B. D. Natarajan *et al.*, "Predicting intracranial progression following stereotactic radiosurgery for brain metastases: Implications for post SRS imaging," *J Radiosurg SBRT*, vol. 6, no. 3, pp. 179–187, 2019.